\documentclass[conference]{IEEEtran}
\usepackage{graphicx}
\usepackage{algorithm}
\usepackage{algorithmic}
\usepackage{amsmath}
\usepackage{multirow}
\usepackage{amssymb}
\usepackage{hyperref}
\usepackage{booktabs}
\usepackage{colortbl,xcolor} 
\newcommand{\best}[1]{\cellcolor{gray!20}\textbf{#1}}
\newcommand{\besttwo}[1]{\cellcolor{yellow!20}\textbf{#1}}

\begin{document}
	
	\title{Robust Federated Learning with Confidence-Weighted Filtering and GAN-Based Completion under Noisy and Incomplete Data}
	
    \author{
	\IEEEauthorblockN{
		Alpaslan Gökcen\IEEEauthorrefmark{1}\IEEEauthorrefmark{2}\\
		ORCID: 0000-0002-5164-7109\\
		Ali Boyacı\IEEEauthorrefmark{1}\IEEEauthorrefmark{3}\\
		ORCID: 0000-0002-2553-1911
	}
	\IEEEauthorblockA{\IEEEauthorrefmark{1}Computer Engineering Department, İstanbul Commerce University, Istanbul 34840, Turkey}
	\IEEEauthorblockA{\IEEEauthorrefmark{2}Turkcell Teknoloji, Maltepe 34854, Istanbul, Turkey}
	\IEEEauthorblockA{\IEEEauthorrefmark{3}Grid Communications and Security Group, Electrification and Energy Infrastructures Division,\\ Oak Ridge National Laboratory, Oak Ridge, TN 37830, USA}
	\IEEEauthorblockA{Corresponding Author: alpaslan.gokcen@turkcell.com.tr}
}
	
	\maketitle
	
	\begin{abstract}
		Federated learning (FL) presents an effective solution for collaborative model training while maintaining data privacy across decentralized client datasets. However, data quality issues such as noisy labels, missing classes, and imbalanced distributions significantly challenge its effectiveness. This study proposes a federated learning methodology that systematically addresses data quality issues, including noise, class imbalance, and missing labels. The proposed approach systematically enhances data integrity through adaptive noise cleaning, collaborative conditional GAN-based synthetic data generation, and robust federated model training. Experimental evaluations conducted on benchmark datasets (MNIST and Fashion-MNIST) demonstrate significant improvements in federated model performance, particularly macro-F1 Score, under varying noise and class imbalance conditions. Additionally, the proposed framework carefully balances computational feasibility and substantial performance gains, ensuring practicality for resource constrained edge devices while rigorously maintaining data privacy. Our results indicate that this method effectively mitigates common data quality challenges, providing a robust, scalable, and privacy compliant solution suitable for diverse real-world federated learning scenarios.
	\end{abstract}
	
	\section{Introduction}
	Federated Learning (FL) has emerged as a transformative approach to collaborative machine learning, enabling multiple clients to jointly train a global model without sharing their local data~\cite{McMahan2016CommunicationEfficientLO}. This decentralized framework addresses significant privacy and data governance concerns, particularly in sensitive domains such as healthcare~\cite{Teo2024}, financial services~\cite{liu2023efficientsecurefederatedlearning}, mobile computing~\cite{jere2020federatedlearningmobileedge} , and connected vehicles or autonomous cars~\cite{electronics13142820}. However, despite these advantages, federated learning systems are challenged by severe data quality issues due to the inherently noisy, imbalanced, and incomplete nature of real-world data collected by distributed clients~\cite{XIA2021100008, Li_2020, Bhanbhro2024}.
	
	One common issue in FL environments is label noise, often arising from annotation errors, data corruption, or adversarial interference. Such noisy labels can mislead model training, reducing overall model accuracy and impairing generalization capabilities. Additionally, the heterogeneous distribution of data across clients frequently leads to missing class scenarios, where certain clients possess incomplete or partial class representations. This non-IID (non-independent and identically distributed) data distribution exacerbates training instability and negatively impacts the convergence and performance of global models trained via standard federated averaging approaches.
	
	To tackle these pervasive challenges, such as label noise, missing classes, and class imbalance in federated learning environments, we propose a three stage methodology that systematically improves data quality and model robustness:
	
	Local Noise Cleaning: Each client applies a confidence weighted filtering mechanism to identify and remove mislabeled samples from its local dataset. This process uses a combination of entropy, margin, and clustering based confidence scores, along with adaptive thresholding to retain high quality data.
	
	Federated Conditional GAN Training: Clients collaboratively train lightweight conditional GANs (cGANs) using their cleaned datasets. The training process follows a federated averaging protocol where only model weights are shared, preserving privacy.
	
	Data Completion via Synthetic Generation \& Federated Classifier Training: Clients that lack certain classes generate synthetic samples for those classes using the trained cGAN generator. These generated samples are directly added to local datasets to balance class distributions, without requiring manual human validation. With completed and balanced datasets, all clients participate in the training of a global CNN classifier using either FedAvg or FedProx. This final stage ensures robust model convergence and improved generalization across non-IID client data.

	Our proposed method significantly improves federated learning performance under realistic data quality conditions. We empirically validate the approach using benchmark datasets such as MNIST~\cite{lecun2010mnist} and Fashion-MNIST~\cite{DBLP:journals/corr/abs-1708-07747}, simulating diverse scenarios of label noise and missing classes. Comparative evaluations demonstrate substantial enhancements in data quality metrics and classification F1-Score relative to standard federated learning baselines. Furthermore, the method is designed to remain computationally feasible for resource constrained edge devices, incorporating differential privacy mechanisms to maintain rigorous data protection standards.
	
	In summary, this paper makes the following key contributions:
	
	1. Proposes a comprehensive, three stage federated learning pipeline specifically designed to address noisy labels, missing classes, and imbalanced data distributions.
	2. Introduces a federated collaborative GAN training strategy coupled with adaptive confidence based data cleaning to systematically enhance data quality.
	3. Validates the proposed approach through extensive experiments, showcasing improvements in F1-Score, stability, and robustness against common federated learning challenges.
	
	The remainder of this paper is structured as follows: Section 2 reviews related work and background concepts. Section 3 details our proposed methodology, including noise cleaning, collaborative GAN training, and data completion strategies. Section 4 describes the experimental setup and evaluation metrics, followed by Section 5, which presents and analyzes the results. Section 6 discusses practical considerations for real-world deployment. Finally, Sections 7 and 8 outline limitations, future directions, and summarize our conclusions.

	\section{Background and Related Work}
	Federated Learning (FL) has emerged as a promising paradigm for collaborative model training across decentralized clients while preserving data privacy. However, real-world FL deployments are far from ideal; they are often plagued by noisy labels, non-IID data distributions, missing class samples, and client heterogeneity in both data quality and model architecture. These challenges significantly degrade model performance and hinder convergence. As a result, a growing body of research has sought to improve the robustness of FL systems through various strategies, including noise resilient loss functions, client reliability estimation, adaptive aggregation mechanisms, and synthetic data generation using generative models. In this section, we review recent efforts that address one or more of these challenges, highlighting their contributions, limitations, and relevance to our proposed approach.
	
	Zhao et al. investigate the statistical challenges posed by non-IID data in federated learning and provide a formal analysis of its impact on model convergence and accuracy. The authors demonstrate that the performance of the FedAvg algorithm deteriorates significantly under highly skewed client distributions up to 55\% accuracy loss on keyword spotting tasks compared to IID baselines. They introduce the concept of weight divergence as a proxy for learning degradation and show that it correlates strongly with the earth mover’s distance (EMD) between local and global class distributions. As a mitigation strategy, the paper proposes distributing a small globally shared dataset to all clients, which reduces EMD and improves performance. Experimental results show that even with as little as 5\% shared data, accuracy can improve by up to 30\% on CIFAR-10. This work provides both a theoretical and practical foundation for understanding and addressing distributional imbalance in federated optimization~\cite{https://doi.org/10.48550/arxiv.1806.00582}.

	Augenstein et al. propose the use of generative models to support model development and debugging in federated learning (FL) settings where direct access to raw data is restricted due to privacy concerns. Their work demonstrates that differentially private generative models specifically RNNs for text and GANs for images can effectively simulate representative data samples, enabling practitioners to identify common data issues such as label noise, misclassifications, and underrepresented classes. The study introduces a novel framework that integrates federated learning with user level differential privacy to train these generative models without compromising individual data privacy. Experimental results show that synthetic data generated by these models can serve as a proxy for direct data inspection, offering practical solutions for debugging and bias detection in decentralized and privacy sensitive environments. This approach highlights the value of generative modeling as a tool for enhancing robustness in FL workflows, particularly under constraints of data inaccessibility~\cite{augenstein2020generativemodelseffectiveml}.
	
	Yang et al. address the challenge of noisy labels in federated learning (FL) settings, where decentralized data annotations often vary in quality due to differences in clients' labeling processes or background knowledge. Their approach proposes a robust FL framework that mitigates label noise by interchanging class wise feature centroids between the server and clients. This centroid based coordination helps align the decision boundaries of local models despite differing noise distributions, thereby reducing weight divergence during model aggregation. Additionally, they introduce a confidence based sample selection strategy, where only low-loss (i.e., likely correct) instances are used in training, and noisy labels are corrected via a global guided pseudo labeling mechanism leveraging the central model. Experimental results on CIFAR-10 and Clothing1M demonstrate that their method consistently outperforms existing baselines, particularly under varying levels and distributions of label noise. This study highlights the importance of structure aware feature coordination and pseudo label correction to ensure robust learning in noisy FL environments~\cite{9713942}.
	
	Wu et al. introduce FedCG, a federated learning framework designed to balance privacy protection and model performance through the use of conditional generative adversarial networks (cGANs). In their approach, each client decomposes its model into a private extractor and a public classifier, retaining the extractor locally while sharing only the generator and classifier with the server. This architectural design mitigates the risk of gradient based privacy attacks, such as Deep Leakage from Gradients (DLG), by ensuring that components exposed to the server do not directly process raw data. The global generator and classifier are constructed on the server via knowledge distillation from client shared generators and classifiers, eliminating the need for public datasets. Extensive experiments across both IID and non-IID scenarios demonstrate that FedCG maintains competitive accuracy while significantly improving privacy preserving capabilities compared to traditional FL baselines like FedAvg, FedProx, and FedSplit. This work highlights the utility of conditional GANs in federated settings for privacy preserving knowledge sharing and personalized local model enhancement~\cite{Wu_2022}.
	
	Gupta et al. propose FedAR+, a federated learning framework tailored for appliance recognition in smart residential environments, particularly under the dual challenges of data privacy and noisy labels. The method enables decentralized model training across clients without sharing raw power consumption data, thereby preserving privacy. To address mislabeled data, the authors introduce an adaptive noise handling mechanism based on a joint loss function that incorporates label distributions and weight parameters. This allows the model to iteratively refine label estimates while simultaneously updating network weights. Furthermore, a custom aggregation function is employed to mitigate biases arising from non-IID client data distributions. Experimental results across multiple datasets including a real-world smart plug dataset demonstrate that FedAR+ can maintain high recognition accuracy (over 85\%) even when up to 30\% of training labels are noisy. This work underscores the potential of federated learning frameworks to deliver robust, privacy preserving models in real-world IoT scenarios, especially when dealing with unreliable supervision~\cite{10.1145/3576841.3585921}.
	
	Wu et al. present FEDCNI, a federated learning framework designed to address the joint challenges of label noise and class imbalance in non-IID client data without relying on clean proxy datasets. The proposed system consists of a noise resilient local solver and a robust global aggregator. At the client level, it introduces a prototypical noise detection mechanism that leverages cosine similarity and Gaussian Mixture Models to differentiate between clean and noisy samples, followed by curriculum based pseudo labeling and a denoise Mixup strategy to mitigate the impact of incorrect annotations. On the server side, FEDCNI adopts a switching re-weighted aggregation strategy, dynamically adjusting the importance of local updates based on the learning stage and estimated noise levels. Experimental evaluations across CIFAR-10, CIFAR-100, and Clothing1M datasets demonstrate that FEDCNI achieves state-of-the-art performance under both synthetic and natural label noise, often rivaling or surpassing clean data baselines. This work highlights the importance of tailored local noise handling and adaptive aggregation for robust federated learning in realistic, heterogeneous environments\cite{10219725}.
	
	Wu et al. introduce FedNoRo, a two stage federated learning framework designed to address real-world challenges arising from class imbalance and heterogeneous label noise. Unlike prior approaches that assume globally balanced data, FedNoRo models a more realistic setting where the distribution of classes and noise rates varies across clients. In the first stage, noisy clients are identified using per class average loss indicators and a Gaussian Mixture Model, ensuring privacy by transmitting only statistical loss summaries. In the second stage, the framework employs differentiated training strategies: clean clients use cross entropy loss, while noisy clients utilize knowledge distillation to reduce the impact of corrupted labels. Additionally, a distance aware aggregation mechanism is applied to minimize the influence of noisy client updates during model aggregation. Evaluations on medical datasets (ICH and ISIC 2019) demonstrate that FedNoRo outperforms existing methods under both label noise and class imbalance, offering a robust and privacy conscious solution for federated learning in practical scenarios~\cite{10.24963/ijcai.2023/492}.
	
	Liang et al. propose FedNoisy, the first comprehensive benchmark specifically designed to evaluate federated learning under noisy label conditions. The authors develop a standardized simulation pipeline encompassing 20 federated settings across six datasets with both synthetic and real-world label noise. FedNoisy systematically examines the impact of heterogeneous data distributions and diverse noise types including symmetric, asymmetric, and real-world noise under various IID and non-IID partitioning schemes. In addition, it incorporates nine baseline algorithms from both centralized noisy label learning (CNLL) and federated learning domains, offering a unified framework for fair and reproducible evaluations. The benchmark highlights critical findings such as the increased difficulty of localized label noise in non-IID environments, the interplay between noise severity and class imbalance, and the non monotonic effects of noise ratio on FL performance. By enabling fine grained evaluations and offering extensible code resources, FedNoisy serves as a foundational tool for advancing robust and noise resilient federated learning methods~\cite{liang2025fednoisyfederatednoisylabel}.

	Li et al. introduce FedNS, a plugin noise aware aggregation strategy for federated learning designed to mitigate the detrimental impact of noisy client data in the input space. Unlike most existing approaches that focus on label noise, FedNS targets real-world input corruptions such as visual distortions or synthetic patch based noise that commonly arise in decentralized environments. The method leverages the gradient norm behavior of local models during early training rounds to identify noisy clients via a single interaction clustering mechanism, thereby preserving privacy. Subsequently, a noise sensitive aggregation strategy is employed to dynamically reweight model updates, assigning greater influence to cleaner clients. FedNS integrates seamlessly with various standard FL algorithms, including FedAvg, FedProx, FedTrimmedAvg, and FedNova. Empirical results across six benchmark datasets and multiple noise types demonstrate substantial improvements in generalization, particularly under high noise severity and non-IID settings. This work broadens the scope of federated robustness research by addressing previously underexplored challenges posed by input level noise and heterogeneous data quality in practical FL deployments~\cite{li2024collaboratively}.
	
	Morafah et al. propose ClipFL, a federated learning framework that addresses the challenge of noisy labels by identifying and excluding low quality clients rather than attempting to correct noisy samples. The method introduces a novel three phase approach: (1) a preclient pruning phase that uses a clean validation set to rank client performance and compute a Noise Candidacy Score (NCS), (2) a client pruning stage that excludes clients with high NCS, and (3) a post-client pruning stage in which standard FL is performed with the remaining clean clients. Experimental results on CIFAR-10 and CIFAR-100 datasets under both IID and non-IID settings demonstrate that ClipFL significantly outperforms baseline FL optimizers and state-of-the-art noise robust methods in terms of accuracy, convergence speed, and communication efficiency. Unlike label correction based approaches that rely on well performing global models, ClipFL eliminates the source of noise at the client level, offering a scalable and efficient alternative for robust federated learning in noisy environments~\cite{10.1145/3706058}.

	Wang et al. propose FedeAMC, a federated learning framework for automatic modulation classification (AMC) that addresses privacy concerns, class imbalance, and varying noise conditions in wireless communication systems. Traditional AMC methods, whether feature based or centralized deep learning based (CentAMC), require extensive labeled data collected from clients, introducing significant privacy risks. In contrast, FedeAMC enables decentralized training on IQ samples at the client level while exchanging only gradients or model weights with the server. To handle class imbalance among clients, the authors integrate balanced cross entropy (BCE) as a loss function, and explore two optimization strategies synchronous stochastic gradient descent (SSGD) and model averaging (MA). Simulation results demonstrate that FedeAMC achieves competitive performance with CentAMC, incurring less than 2\% accuracy loss while significantly enhancing privacy protection. Moreover, the use of BCE accelerates convergence and improves classification performance, particularly under heterogeneous and imbalanced data conditions. This work underscores the efficacy of federated approaches in maintaining model accuracy while ensuring data confidentiality in realistic wireless environments~\cite{9456904}.

	Fang and Ye introduce RHFL, a novel federated learning framework specifically designed to address the dual challenges of label noise and client model heterogeneity. Unlike conventional FL approaches that assume homogeneous client architectures and clean data, RHFL enables decentralized learning among clients with distinct local models and varying noise rates. The method integrates three core components: (1) Knowledge distribution alignment using public datasets to facilitate communication across heterogeneous models without relying on a shared global model; (2) Symmetric loss (SL) to mitigate overfitting to noisy labels by combining cross entropy and reverse cross entropy during local training; and (3) Client Confidence Re-weighting (CCR), a mechanism that quantifies label quality and learning efficiency to reduce the influence of unreliable clients in global aggregation. Experimental results across various noise types and architectures demonstrate that RHFL consistently outperforms baseline methods in both heterogeneous and homogeneous FL scenarios. This work broadens the scope of robust FL by addressing realistic deployment issues such as model heterogeneity, unbalanced data quality, and communication noise~\cite{9878489}.
	
	Jeong et al. propose FedMatch, a federated semi-supervised learning (FSSL) framework designed to handle scenarios in which clients possess partially labeled or entirely unlabeled data. Recognizing the impracticality of assuming fully labeled datasets in real-world FL deployments, FedMatch introduces two complementary innovations: an inter client consistency loss that promotes agreement across distributed models, and a parameter decomposition strategy that isolates supervised and unsupervised learning processes to mitigate inter-task interference. This design allows the method to adapt to both "labels at client" and "labels at server" scenarios, improving training stability and generalization performance. Extensive experiments across IID, non-IID, and streaming data tasks demonstrate that FedMatch consistently outperforms traditional semi-supervised learning baselines (e.g., FixMatch, UDA) when integrated with FL frameworks like FedAvg and FedProx. Moreover, it significantly reduces communication costs by leveraging sparse parameter updates. FedMatch effectively addresses realistic challenges in FL settings, such as partial labeling, non-IID distributions, and high communication overhead, offering a scalable and robust solution for federated learning under limited supervision~\cite{jeong2021federated}.
	
	Zhang et al. address a significant limitation in federated learning (FL) by introducing FedAlign, a framework tailored for settings where clients possess non-identical and even disjoint class labels a scenario referred to as client exclusive classes. Unlike traditional FL methods that assume a consistent class set across clients, FedAlign introduces a two branch architecture comprising a data encoder and a label encoder, and leverages natural language class names as shared semantic anchors. This approach enables clients to align their latent spaces despite working with disjoint class sets. Furthermore, FedAlign incorporates a knowledge distillation mechanism that annotates data for locally unaware classes using semantic similarity and distills this pseudo knowledge into local models. Experimental results on behavioral recognition, medical diagnosis, activity recognition, and text classification datasets demonstrate that FedAlign outperforms existing FL baselines under both single label and multi label classification settings. This work highlights the importance of semantic alignment and distillation in achieving robust global models under severe class heterogeneity~\cite{zhang}.
	
	While prior work has explored noise robust optimization, synthetic sample generation, client pruning, and aggregation strategies, most existing methods address only a subset of FL’s practical challenges. Some focus narrowly on label noise or assume homogeneous model architectures, while others overlook missing classes or the joint effect of noise and non-IID distributions. In contrast, our work proposes a comprehensive and modular framework that integrates multi metric confidence estimation, adaptive filtering, confidence weighted aggregation, class conditional generative modeling, and robust federated optimization via FedProx~\cite{li2020federatedoptimizationheterogeneousnetworks}. By addressing these issues holistically, our method advances the state of robust federated learning under real-world conditions.
	
	Unlike these partial solutions, our proposed method addresses the joint challenges of label noise, missing classes, and non-IID distributions within a unified and modular FL framework.
		
	\section{Proposed Methodology}
	\subsection{Overview}
	To systematically address data quality issues in federated learning (FL), we introduce a comprehensive three stage methodology designed to handle noisy labels, class imbalance, and missing classes effectively. This approach enhances the integrity and representativeness of data, thus significantly improving federated model performance.
	
	\begin{figure}
		\centering
		\includegraphics[width=1\linewidth]{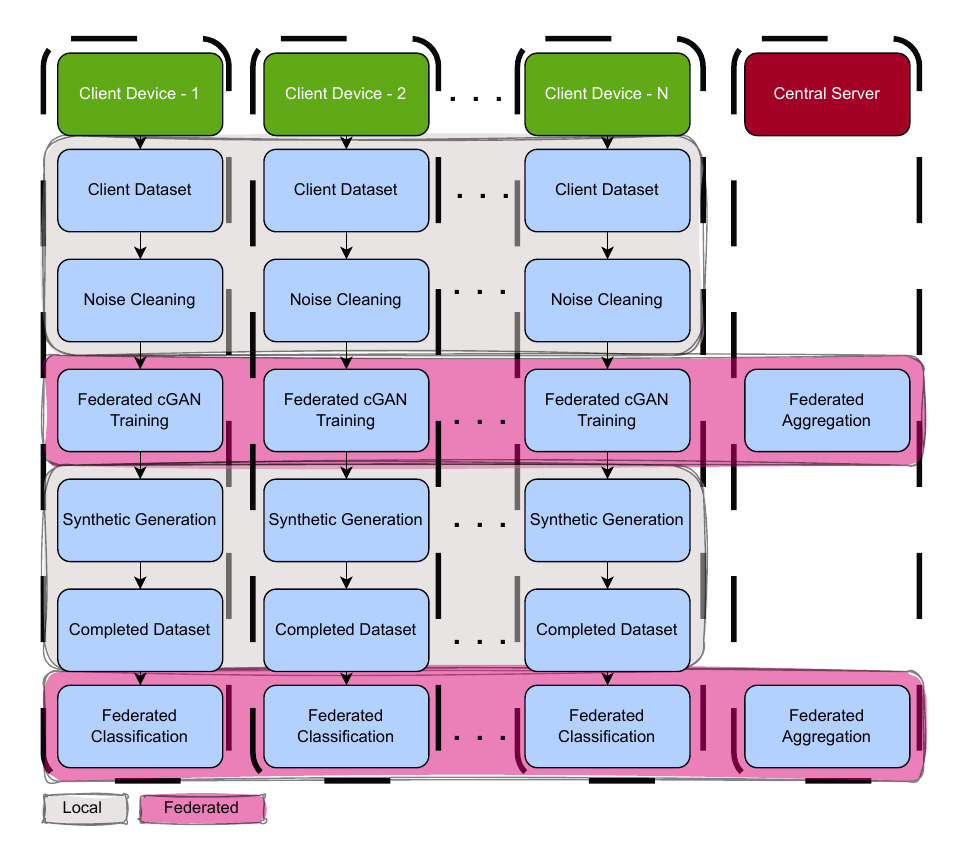}
		\caption{Three-Stage Federated Learning Framework for Robust Training under Noisy and Incomplete Data}
		\label{fig:Framework}
	\end{figure}
	
	Our proposed solution comprises the following sequential stages as shown in Figure \ref{fig:Framework}:
	
	1. Noise Cleaning: Local identification and correction of mislabeled or erroneous samples using advanced ensemble based methods.
	2. Collaborative GAN Training: Federated training of lightweight conditional GANs to generate synthetic data for missing classes.
	3. Data Completion and Federated Training: integration of synthetic data to complete local datasets, followed by federated model training.

	\subsection{Stage 1: Noise Cleaning}
	
	This stage aims to enhance data quality locally at each client through noise detection and correction, ensuring that only high confidence samples are utilized in subsequent federated training rounds. The noise cleaning process involves several substeps:
	
	\begin{itemize}
		\item Stratified K-Fold Cross-Validated CNN: Each client trains a lightweight convolutional neural network (CNN)~\cite{Alzubaidi2021} using stratified K-fold cross-validation~\cite{Kohavi1995ASO} on local data.
		\item Confidence Scoring Metrics: Instances are evaluated using three complementary confidence metrics:
		
		\begin{itemize}
			\item Entropy based confidence: Lower entropy indicates higher certainty in predictions~\cite{kim}.
			\item Margin based confidence: Measures the gap between top two predicted probabilities.
			\item Cluster based confidence: Employs K-means~\cite{MacQueen1967SomeMF, Lloyd1982LeastSQ} clustering on feature embeddings and calculates Silhouette scores to detect inconsistent samples.
		\end{itemize}
		
		\item Adaptive Thresholding: Confidence scores are combined to determine adaptive, client specific thresholds. Samples below these thresholds are marked as noisy and removed.
		\item Ensemble Aggregation: Final noise cleaned datasets are created by aggregating models trained on multiple data folds, weighted by confidence scores to further enhance robustness.
		
	\end{itemize}
	
	\begin{algorithm}[H]
		\caption{Client Level Confidence Based Cleaning}
		\label{alg:client_cleaning}
		\begin{algorithmic}[1]
			\REQUIRE Client dataset $(x, y)$, number of folds $K$
			\ENSURE Cleaned dataset $(x_{\text{clean}}, y_{\text{clean}})$ and final model $M$
			
			\STATE Split dataset into $K$ stratified folds
			\FOR{each fold $k = 1$ to $K$}
			\STATE Train CNN model $M_k$ on $(K-1)$ folds
			\FOR{each sample $x_i$ in fold $k$}
			\STATE Compute prediction probabilities $p = M_k(x_i)$
			\STATE Compute entropy: $C_{\text{ent}}(x_i) = -\sum_c p_c \log p_c$
			\STATE Compute margin: $C_{\text{margin}}(x_i) = p_{1st} - p_{2nd}$
			\STATE Compute cluster score $C_{\text{cluster}}(x_i)$ using silhouette on $[x_i, p]$
			\ENDFOR
			\ENDFOR
			
			\STATE Calculate aggregated confidence:
			\[
			C_{\text{agg}}(x_i) = \frac{1}{3} \left( C_{\text{ent}}(x_i) + C_{\text{margin}}(x_i) + C_{\text{cluster}}(x_i) \right)
			\]
			
			\STATE Determine adaptive threshold $T$ using mean, median, and 75th percentile of scores
			\STATE Initialize $D_{\text{clean}} = \emptyset$
			
			\FOR{each sample $(x_i, y_i)$ in $(x, y)$}
			\IF{$C_{\text{agg}}(x_i) \geq T$}
			\STATE Add $(x_i, \arg\max p)$ to $D_{\text{clean}}$
			\ENDIF
			\ENDFOR
			
			\STATE Train final model $M$ on $D_{\text{clean}}$
			\RETURN $D_{\text{clean}}, M$
		\end{algorithmic}
	\end{algorithm}
	
	As illustrated in Algorithm~\ref{alg:client_cleaning}, the proposed procedure aims to identify and retain high confidence data samples from a potentially noisy local client dataset by leveraging multiple confidence estimation strategies and an adaptive thresholding mechanism.
	
	The algorithm starts by receiving a local dataset $D = \{(x_i, y_i)\}_{i=1}^N$ and a predefined number of folds $K$ as input. The dataset is partitioned into $K$ stratified folds to perform cross validation. For each fold $k$, a CNN model $M_k$ is trained on the remaining $K-1$ folds, ensuring that the validation data in each fold is never seen during training.
	
	For every validation sample $x_i$ in fold $k$, the trained model $M_k$ generates a probability distribution $p = M_k(x_i)$ over the class labels. Based on these predictions, three types of confidence scores are computed:
	
	\begin{itemize}
		\item \textbf{Entropy based confidence} quantifies uncertainty in predictions using the formula $C_{\text{ent}}(x_i) = -\sum_c p_c \log p_c$. Lower entropy indicates higher confidence.
		
		\item \textbf{Margin based confidence} is calculated as the difference between the top two predicted probabilities, i.e., $C_{\text{margin}}(x_i) = p_{1st} - p_{2nd}$, where $p_{1st}$ and $p_{2nd}$ are the highest and second highest values in $p$.
		
		\item \textbf{Cluster based confidence} is derived by performing K-means clustering on the joint space of input features and predicted probabilities. Silhouette scores are computed to assess how well each sample fits within its assigned cluster, resulting in the cluster based confidence $C_{\text{cluster}}(x_i)$.
	\end{itemize}
	
	These three confidence scores are then aggregated for each sample using a simple average, as shown below in Equation \ref{eq:confidence_agg}:

	\begin{equation}
		C_{\text{agg}}(x_i) = \frac{1}{3} \left( C_{\text{ent}}(x_i) + C_{\text{margin}}(x_i) + C_{\text{cluster}}(x_i) \right)
		\label{eq:confidence_agg}
	\end{equation}

	In Equation \ref{eq:adaptive_th} following this, an adaptive threshold $T$ is determined based on the distribution of aggregated confidence scores. Specifically, the threshold is computed as the average of the mean, median, and 75th percentile:
	\begin{equation}
		T = \frac{1}{3} \left( \text{mean}(C) + \text{median}(C) + P_{75}(C) \right)
		\label{eq:adaptive_th}
	\end{equation}
	
	All samples whose confidence scores satisfy $C_{\text{agg}}(x_i) \geq T$ are selected as trustworthy. For each selected sample, the predicted label is obtained using the $\arg\max$ of the probability vector $p$, and the resulting pair $(x_i, \arg\max p)$ is added to the cleaned dataset $D_{\text{clean}}$.
	
	After filtering, a final CNN model $M$ is trained on the clean dataset $D_{\text{clean}}$. This model is expected to exhibit improved generalization performance due to the exclusion of noisy or ambiguous samples during training.
	
	Algorithm~\ref{alg:client_cleaning} presents a comprehensive pipeline for local noise reduction that combines model confidence estimation, unsupervised clustering, and adaptive decision boundaries to isolate high quality data under federated learning settings or other decentralized scenarios.

	\subsection{Stage 2: Collaborative GAN Training}
	To address missing class issues, clients collaboratively train lightweight conditional GANs (cGANs) using a federated averaging approach. This collaborative training ensures high quality synthetic data generation while preserving privacy constraints:
	\begin{itemize}
		\item Conditional GAN~\cite{mirza2014conditionalgenerativeadversarialnets} Architecture: A lightweight class conditional GAN architecture is adopted, suitable for edge device computational constraints, enabling controlled generation of class specific samples.
		\item Federated Averaging (FedAvg)~\cite{McMahan2016CommunicationEfficientLO} of GAN Parameters: Clients train local GAN instances and periodically synchronize their generator and discriminator parameters with a central server through FedAvg, ensuring privacy by exchanging only model parameters rather than raw data.
		\item Differential Privacy: Calibrated differential privacy mechanisms are integrated during parameter aggregation to provide rigorous privacy guarantees, protecting against inference attacks.
	\begin{algorithm}[H]
		\caption{Federated GAN Training Across Clients}
		\label{alg:federated_training}
		\begin{algorithmic}[1]
			\REQUIRE Set of cleaned clients $\mathcal{C}$, number of epochs $E$, regularization coefficient $\mu$
			
			\FOR{each epoch $e = 1$ to $E$}
			\STATE Initialize epoch losses: $g_{\text{epoch}} \gets 0$, $d_{\text{epoch}} \gets 0$
			\STATE Initialize empty weight lists: $\mathcal{W}_G = [\,]$, $\mathcal{W}_D = [\,]$
			
			\STATE Select global models $G^{(global)}$, $D^{(global)}$ from any client in $\mathcal{C}$
			
			\STATE \textbf{Parallel client training:}
			\FORALL{clients $c_i \in \mathcal{C}$ \textbf{in parallel}}
			\STATE $(g_{\text{loss}}, d_{\text{loss}}, (\theta_G^i, \theta_D^i)) \gets$ \texttt{train\_one\_epoch}$(G^{(global)}, D^{(global)}, \mu)$
			\STATE $g_{\text{epoch}} \mathrel{+}= g_{\text{loss}}$, $d_{\text{epoch}} \mathrel{+}= d_{\text{loss}}$
			\STATE Append $\theta_G^i$ to $\mathcal{W}_G$, and $\theta_D^i$ to $\mathcal{W}_D$
			\ENDFOR
			
			\STATE Average weights:
			\[
			\theta_G^{(global)} = \frac{1}{|\mathcal{C}|} \sum_i \theta_G^i, \quad
			\theta_D^{(global)} = \frac{1}{|\mathcal{C}|} \sum_i \theta_D^i
			\]
			
			\STATE Update all clients with global weights:
			\FOR{each client $c_i \in \mathcal{C}$}
			\STATE $c_i.\texttt{set\_weights}(\theta_G^{(global)}, \theta_D^{(global)})$
			\ENDFOR
			
			\STATE Compute average losses:
			\[
			\bar{g}_{\text{loss}} = \frac{g_{\text{epoch}}}{|\mathcal{C}|}, \quad
			\bar{d}_{\text{loss}} = \frac{d_{\text{epoch}}}{|\mathcal{C}|}
			\]
			
			\ENDFOR
		\end{algorithmic}
	\end{algorithm}
	
	As illustrated in Algorithm~\ref{alg:federated_training}, the training process consists of federated optimization for a conditional Generative Adversarial Network (GAN), where multiple clients train their local generator and discriminator models and collaboratively update shared global models.
	
	At the beginning of each global epoch $e$, two accumulators are initialized to store the generator and discriminator losses: $g_{\text{epoch}}$ and $d_{\text{epoch}}$. In addition, two lists $\mathcal{W}_G$ and $\mathcal{W}_D$ are created to store the local model weights from each client after one round of training.
	
	The global generator $G^{(global)}$ and discriminator $D^{(global)}$ are cloned from any participating client (e.g., the first client in the list). These serve as the initialization point for all clients during the current communication round.
	
	Clients then enter a parallel training phase, where each client $c_i \in \mathcal{C}$ invokes its local \texttt{train\_one\_epoch} function using the current global models as inputs and a regularization coefficient $\mu$. This function returns three values: the local generator loss $g_{\text{loss}}$, the local discriminator loss $d_{\text{loss}}$, and the updated weights $(\theta_G^i, \theta_D^i)$. As these results are collected, the global epoch loss accumulators and model weight lists are updated accordingly.
	
	Once all clients complete local training, their respective model weights are aggregated. Specifically, the global generator and discriminator weights are computed by taking the average over all clients as shown in Equation \ref{eq:generator_agg}:
	\begin{equation}
	\theta_G^{(global)} = \frac{1}{|\mathcal{C}|} \sum_i \theta_G^i, \quad \theta_D^{(global)} = \frac{1}{|\mathcal{C}|} \sum_i \theta_D^i
	\label{eq:generator_agg}
	\end{equation}
	
	These aggregated weights are then sent back to all clients to synchronize their local models with the global ones. This ensures that all clients begin the next communication round from a consistent and jointly optimized state.
	
	Following the weight update in Equation \ref{eq:generator_loss}, the algorithm computes the average generator and discriminator losses across all clients:
	\begin{equation}
	\bar{g}_{\text{loss}} = \frac{g_{\text{epoch}}}{|\mathcal{C}|}, \quad \bar{d}_{\text{loss}} = \frac{d_{\text{epoch}}}{|\mathcal{C}|}
	\label{eq:generator_loss}
	\end{equation}

	\end{itemize}
	
	\subsection{Stage 3: Data Completion and FL Training}
	The final stage leverages the trained collaborative GANs to complete local datasets with synthetic samples for missing or underrepresented classes, followed by robust federated training:
	\begin{itemize}
		\item Synthetic Data Generation: Clients use the globally aggregated GAN generators to produce synthetic samples for missing classes, thereby balancing local datasets.
		\item Centralized Validation Classifier: A centralized classifier, pretrained or federatively trained using balanced data, validates generated samples. Class balanced loss functions and adaptive thresholds ensure unbiased and semantically accurate synthetic data.
		\item Dataset Completion: Validated synthetic samples are integrated into local datasets, completing class coverage and enhancing representativeness.
		\item Robust Federated Model Training: The enhanced datasets are used to train global models via standard or regularized aggregation (e.g., FedAvg, FedProx), improving stability and performance, particularly under non-IID and noisy conditions.
		
	\end{itemize}

	\begin{algorithm}[H]
		\caption{Generate Samples for Missing Classes}
		\label{alg:generate_missing}
		\begin{algorithmic}[1]
			\REQUIRE Set of missing classes $\mathcal{M}$, generator model $G$, sample size $s$ per class
			\ENSURE Generated dataset $(x_{\text{gen}}, y_{\text{gen}})$
			
			\STATE Initialize empty lists: $x_{\text{gen}} \gets [\,]$, $y_{\text{gen}} \gets [\,]$
			
			\FOR{each class label $c \in \mathcal{M}$}
			\STATE Sample latent vectors: $z \sim \mathcal{N}(0, I)^{s \times 100}$
			\STATE Create label vector: $y = [c, c, \ldots, c] \in \mathbb{Z}^s$
			\STATE Generate images: $\hat{x} = G(z, y)$
			\STATE Normalize: $\hat{x} \leftarrow (\hat{x} + 1)/2$
			\STATE Append $\hat{x}$ to $x_{\text{gen}}$, $y$ to $y_{\text{gen}}$
			\ENDFOR
			
			\STATE Concatenate all generated samples and labels
			
			\RETURN $(x_{\text{gen}}, y_{\text{gen}})$
		\end{algorithmic}
	\end{algorithm}
	
	As shown in Algorithm~\ref{alg:generate_missing}, this procedure is designed to synthetically generate labeled data for classes that are missing or underrepresented on the client side in a federated learning setting. The approach leverages a pretrained conditional generator $G$ to produce data conditioned on specific class labels.
	
	The algorithm begins by receiving three key inputs: the set of missing class labels $\mathcal{M}$, a generator model $G$, and a target sample size $s$ per class. For each class $c \in \mathcal{M}$, the generator is queried to produce $s$ synthetic images.
	
	To achieve this, a latent input matrix $z \in \mathbb{R}^{s \times 100}$ is sampled from a standard multivariate Gaussian distribution. In parallel, a label vector $y \in \mathbb{Z}^s$ is created where each entry is set to $c$, indicating the desired class for all generated samples.
	
	The conditional generator $G$ then takes $z$ and $y$ as inputs and produces a set of synthetic samples $\hat{x} = G(z, y)$. These generated images typically lie in the range $[-1, 1]$ due to the use of a \texttt{tanh} activation in the output layer. Therefore, the outputs are linearly transformed to the range $[0, 1]$ via the normalization as shown in Equation \ref{eq:normalization}:
	\begin{equation}
		\hat{x} \leftarrow \frac{\hat{x} + 1}{2}
		\label{eq:normalization}
	\end{equation}

	This normalization ensures that the generated images are compatible with downstream models trained on normalized real-world data. After processing each class in $\mathcal{M}$, the generated samples and their corresponding labels are concatenated into two arrays: $x_{\text{gen}}$ and $y_{\text{gen}}$.
	
	The output of Algorithm~\ref{alg:generate_missing} is a fully labeled synthetic dataset that can be used to augment training data on clients that lack examples for certain classes. This is particularly valuable in non-IID federated learning environments where class imbalance and data heterogeneity can significantly impact model performance. By enriching local datasets with synthetic samples, the algorithm helps to mitigate class missing scenarios and improve generalization during federated training.
	
	\begin{algorithm}[H]
		\caption{Federated Training with FedProx}
		\label{alg:federated_fedprox}
		\begin{algorithmic}[1]
			\REQUIRE Global model $M$, client datasets $\mathcal{D} = \{(x_i, y_i)\}$, total epochs $E$, patience $p$, tolerance $\delta$, regularization coefficient $\mu$
			\ENSURE Trained global model $M$
			
			\STATE Initialize best accuracy $a_{\text{best}} \gets 0$, wait counter $w \gets 0$
			\STATE Get global weights: $w^{(global)} \gets M.\texttt{get\_weights}()$
			
			\FOR{each epoch $e = 1$ to $E$}
			\STATE Initialize list of local weights and sample counts
			\STATE Compute total sample count $N = \sum_i |x_i|$
			
			\FOR{each client $i$ with data $(x_i, y_i)$}
			\STATE Initialize local model $M_i \gets \texttt{build\_cnn\_model}()$
			\STATE Set $M_i$ weights: $w_i \gets w^{(global)}$
			\STATE Define optimizer with exponential learning rate decay
			\FOR{each minibatch $(x_b, y_b) \subset (x_i, y_i)$}
			\STATE Forward pass and compute cross entropy loss $\mathcal{L}_{\text{CE}}$
			\STATE Compute proximal term: $\mathcal{L}_{\text{prox}} = \sum_j \| w_j - w^{(global)}_j \|^2$
			\STATE Total loss: $\mathcal{L} = \mathcal{L}_{\text{CE}} + \frac{\mu}{2} \mathcal{L}_{\text{prox}}$
			\STATE Backpropagate, clip gradients, update weights
			\ENDFOR
			\STATE Append $(w_i, |x_i|)$ to local weight list
			\ENDFOR
			
			\STATE Compute weighted global average:
			\[
			w^{(new)} = \sum_i \frac{|x_i|}{N} \cdot w_i
			\]
			\STATE Update global model: $M \gets w^{(new)}$
			
			\STATE Evaluate $M$ on validation set to obtain accuracy $a$
			\IF{$a > a_{\text{best}} + \delta$}
			\STATE $a_{\text{best}} \gets a$, $w \gets 0$
			\ELSE
			\STATE $w \gets w + 1$
			\IF{$w \geq p$}
			\STATE \textbf{break} \COMMENT{Early stopping triggered}
			\ENDIF
			\ENDIF
			\ENDFOR
			
			\RETURN $M$
		\end{algorithmic}
	\end{algorithm}
	
	In Algorithm~\ref{alg:federated_fedprox}, the proposed federated training procedure is based on the FedProx optimization framework and includes support for adaptive early stopping. The algorithm is designed to train a global model across multiple decentralized clients, each of which performs local training with a proximal regularization term to prevent divergence from the global objective.
	
	At the beginning of training, the server initializes the global model weights and tracks the best validation accuracy achieved so far, as well as a patience counter used for early stopping. For each federated epoch, the total number of samples across all clients is computed to enable weighted aggregation later.
	
	Each client $i$ receives the global model weights $w^{(global)}$ and initializes its own local model accordingly. An exponential decay scheduler is used to adjust the learning rate over time. In Equation \ref{eq:crs_loss}, the client then iterates over its local data in mini batches and computes the standard cross entropy loss $\mathcal{L}_{\text{CE}}$ along with a proximal term:
	
	\begin{equation}
			\mathcal{L}_{\text{prox}} = \sum_j \| w_j - w^{(global)}_j \|^2
		\label{eq:crs_loss}
	\end{equation}

	This term penalizes deviations of the local weights from the global weights, thereby stabilizing learning under non-IID data distributions. The total loss is defined as shown in Equation \ref{eq:total_loss}:

	\begin{equation}
		\mathcal{L} = \mathcal{L}_{\text{CE}} + \frac{\mu}{2} \mathcal{L}_{\text{prox}}
		\label{eq:total_loss}
	\end{equation}
	where $\mu$ is a tunable regularization coefficient. Gradients are computed with respect to the total loss, clipped for stability, and applied to update the local model.
	
	Once all clients have completed their local updates, the server performs a weighted federated averaging step to obtain the new global model as fallows in Equation \ref{eq:w_average}:
	\begin{equation}
			w^{(new)} = \sum_i \frac{|x_i|}{N} \cdot w_i
		\label{eq:w_average}
	\end{equation}

	where $|x_i|$ is the number of samples at client $i$, and $N$ is the total number of training samples across all clients.
	
	After updating the global model, it is evaluated on a separate test or validation dataset to compute its current accuracy. If the accuracy has improved by at least $\delta$ compared to the best observed so far, the early stopping counter is reset. Otherwise, the counter is incremented. If the counter exceeds a predefined patience threshold $p$, the training process is terminated early to prevent overfitting and save computational resources.

	\section{Experimental Setup}
	To rigorously evaluate the effectiveness and robustness of our proposed federated learning methodology, we conducted a comprehensive set of experiments using two widely adopted image classification datasets: MNIST and FashionMNIST. These datasets differ significantly in complexity and visual characteristics, allowing us to assess the generalizability of our method across both simple and challenging domains.
	
	Our experimental design systematically varied two key real-world constraints frequently encountered in federated learning environments: \textit{label noise} and \textit{missing class samples}. Controlled noise ratios were injected into the training labels to simulate erroneous annotations, while missing class sizes were introduced by removing specific class labels entirely from individual clients. These configurations enabled us to test the robustness of each model under increasing levels of data corruption and heterogeneity.
	
	All models were evaluated under identical conditions in terms of data partitioning, local training epochs, and communication rounds. Each experiment was repeated on both datasets to highlight the sensitivity of different models to dataset complexity. For statistical reliability, each experiment was conducted 50 times with the same dataset and configuration, and average results were reported.
	
	We present and discuss results from six federated models that combine or isolate three main methodological components: local noise cleaning, GAN based data augmentation, and federated optimization strategy (FedAvg or FedProx). Performance comparisons are made using classification macro-F1 score to ensure a multi dimensional evaluation of both quality and training dynamics.

	\subsection{Datasets and Simulation}
	We conducted experiments using two widely recognized image classification datasets. The MNIST dataset consists of 60,000 grayscale training images and 10,000 test images, representing handwritten digits from 0 to 9. The Fashion-MNIST dataset similarly comprises 60,000 grayscale training images and 10,000 test images, spanning 10 fashion related classes such as shirts, pants, and shoes.
	
	To simulate realistic federated learning challenges, we introduced asymmetric label noise by randomly mislabeling 10\%, 30\%, 50\%, and 70\% of the training samples, assigning them to semantically related but incorrect classes. Additionally, certain classes were randomly removed from subsets of client datasets to create non-IID data distributions. Experiments involved 10 clients, each holding distinct subsets of the original datasets participating in each communication round. We executed up to 50 federated learning rounds using Federated Averaging (FedAvg) to synchronize and aggregate global model parameters.
	
	\subsection{Noise Model}
	
	As shown in Algorithm~\ref{alg:add_noise}, this method injects label noise into a local dataset while explicitly excluding any classes that are considered missing or unavailable. The goal is to preserve the integrity of data corresponding to missing classes, while introducing a controlled level of noise among the remaining valid samples to simulate realistic label corruption scenarios.
	
	The dataset $(x, y)$ is first partitioned into two disjoint subsets:
	\begin{itemize}
		\item A \textbf{valid set} consisting of samples whose labels do not belong to the missing class set $\mathcal{M}$, i.e., $y \notin \mathcal{M}$.
		\item A \textbf{non valid set} consisting of samples labeled with one of the missing classes, i.e., $y \in \mathcal{M}$.
	\end{itemize}
	
	To achieve a desired noise ratio $\rho$, the algorithm computes how many samples must be corrupted and added to the dataset. Two different cases are considered:
	
	\begin{enumerate}
		\item If the non valid set is large enough to provide noise samples without altering the valid set significantly, the required number of noisy samples is computed using as shown in Equation \ref{eq:non_valid_set}:
		\begin{equation}
			n_{\text{noise}} = \left\lfloor \frac{\rho \cdot |y_{\text{valid}}|}{1 - \rho} \right\rfloor
			\label{eq:non_valid_set}
		\end{equation}

		This equation is derived from the definition of noise ratio as shown in Equation \ref{eq:noise_ratio}:
		\begin{equation}
			\rho = \frac{n_{\text{noise}}}{n_{\text{valid}} + n_{\text{noise}}}
			\label{eq:noise_ratio}
		\end{equation}

		\item In Equation \ref{eq:valid_set}, if the non valid set is insufficient, the algorithm adjusts the size of the valid set to ensure that the final dataset reflects the desired ratio. In this case, it solves for $n_{\text{valid}}$ instead, keeping the available noisy samples fixed:

		\begin{equation}
			n_{\text{valid}} = \left\lfloor \frac{|y_{\text{non}}|}{\rho} - |y_{\text{non}}| \right\rfloor
			\label{eq:valid_set}
		\end{equation}
	\end{enumerate}
	
	Once the appropriate number of samples has been determined, new noisy labels are assigned by sampling from the label distribution of the valid set. These are paired with feature vectors from the non valid set. This ensures that the introduced noise is label wise consistent with the distribution of observed (non missing) classes.
	
	Finally, the selected noisy samples are concatenated with a subset of the valid samples to form the output dataset $(x', y')$, which contains a controlled amount of noise while maintaining the exclusion of any missing classes. This approach is particularly important in federated learning scenarios where some clients may lack specific classes and should not introduce misleading labels for classes they have never observed.
	
	\begin{algorithm}[H]
		\caption{Noise Injection Excluding Missing Classes}
		\label{alg:add_noise}
		\begin{algorithmic}[1]
			\REQUIRE Client data $(x, y)$, set of missing classes $\mathcal{M}$, noise ratio $\rho$
			\ENSURE Noisy inputs $x'$, noisy labels $y'$
			
			\STATE Split dataset into:
			\begin{itemize}
				\item Valid set: $(x_{\text{valid}}, y_{\text{valid}}) \gets$ samples where $y \notin \mathcal{M}$
				\item Non valid set: $(x_{\text{non}}, y_{\text{non}}) \gets$ samples where $y \in \mathcal{M}$
			\end{itemize}
			
			\STATE Compute number of noisy samples to generate:
			\[
			n_{\text{noise}} = \left\{
			\begin{array}{ll}
				\left\lfloor \rho \cdot \frac{|y_{\text{valid}}|}{1 - \rho} \right\rfloor & \text{if } \rho < \frac{|y_{\text{non}}|}{|y|} \\\\
				\left\lfloor \frac{|y_{\text{non}}|}{\rho} - |y_{\text{non}}| \right\rfloor & \text{otherwise}
			\end{array}
			\right.
			\]
			
			\STATE Select noise candidates from $x_{\text{non}}$ and assign labels sampled from $y_{\text{valid}}$
			
			\STATE Optionally trim $x_{\text{valid}}$ to maintain desired noise ratio
			
			\STATE Concatenate noisy samples and remaining valid samples:
			\[
			x' = x_{\text{valid}} \cup x_{\text{noise}}, \quad
			y' = y_{\text{valid}} \cup y_{\text{noise}}
			\]
			
			\RETURN $(x', y')$
		\end{algorithmic}
	\end{algorithm}

	\subsection{Baseline Methods}
	To comprehensively assess the advantages of our proposed methodology, we conducted comparisons against several established baseline methods. The first baseline was standard FedAvg, representing a conventional federated learning approach without any data cleaning or synthetic augmentation. The second baseline employed FedProx under the same conditions, serving as a regularized alternative to FedAvg in the absence of data preprocessing or augmentation.

	These two baseline variants allow us to isolate the contribution of the FedProx optimization technique under noisy and incomplete conditions, providing a more detailed understanding of optimization level robustness.

	In this study, we compare six different federated learning models under various noise and missing class conditions. The models include: 
	\textit{CleanAvg}, which combines confidence based data cleaning with the FedAvg algorithm;
	\textit{CleanProx}, which integrates the same cleaning strategy with FedProx optimization;
	\textit{GenCleanAvg}, which additionally augments the cleaned data with conditional GAN generated samples before applying FedAvg;
	\textit{GenCleanProx}, the same but followed by FedProx;
	\textit{FedAvg (Noisy)}, the baseline model trained directly on noisy data using FedAvg;
	and \textit{FedProx (Noisy)}, the corresponding baseline trained with FedProx.

	\begin{table*}[t]
		\centering
		\caption{Model Architectures and Parameter Counts}
		\label{tab:model_specs}
		\begin{tabular}{p{3cm}p{2.2cm}p{11cm}}
			\toprule
			\textbf{Model} & \textbf{Parameters} & \textbf{Architecture Summary} \\
			\midrule
			
			CNN Classifier & $\sim$1.2M &
			Conv2D(32) + BN $\rightarrow$ MaxPool $\rightarrow$ Conv2D(64) + BN $\rightarrow$ MaxPool $\rightarrow$ Flatten $\rightarrow$ Dense(128) + Dropout(0.5) $\rightarrow$ Dense(10, Softmax) \\
			
			Generator (cGAN) & $\sim$590K &
			Linear layers: 110 $\rightarrow$ 256 $\rightarrow$ 512 $\rightarrow$ 1024 $\rightarrow$ 784; LeakyReLU activations; output reshaped to $28{\times}28$ with Tanh \\
			
			Discriminator (cGAN) & $\sim$1.1M &
			Linear layers: 794 $\rightarrow$ 1024 $\rightarrow$ 512 $\rightarrow$ 256 $\rightarrow$ 1; LeakyReLU + Dropout; Sigmoid output \\
			
			FedProx Regularization & $\mu=0.01$ &
			Adds proximal term $\frac{\mu}{2} \sum_j \| w_j - w_j^{(global)} \|^2$ to local loss \\
			\bottomrule
		\end{tabular}
	\end{table*}

	The CNN classifier is built on a lightweight LeNet~\cite{lenet} like structure with batch normalization and dropout, making it effective for small grayscale datasets like MNIST or FashionMNIST. Dropout at 0.5 helps prevent overfitting in client specific settings~\cite{article_dropout} as shown in Table ~\ref{tab:model_specs}.
	
	The architecture includes commonly used deep learning components: Batch Normalization (BN) stabilizes and accelerates training by normalizing layer inputs~\cite{ioffe2015batchnormalizationacceleratingdeep}. ReLU and LeakyReLU are nonlinear activation functions that help mitigate the vanishing gradient problem. Max Pooling (MaxPool) reduces spatial dimensions while retaining salient features. Dropout randomly deactivates neurons during training to prevent overfitting. Tanh and Sigmoid are used in output layers to produce bounded values. Dense (Fully Connected) layers connect all neurons from one layer to the next.

	MaxPool or Max Pooling is employed to downsample feature maps while retaining the most significant features, commonly used in convolutional neural networks. Dropout is a regularization method that randomly sets a fraction of the neurons to zero during training, helping to prevent overfitting. Tanh and Sigmoid are activation functions used in output layers for generating bounded outputs. Finally, Dense or Fully Connected (FC) layers refer to standard neural layers where each neuron is connected to all neurons in the preceding layer.
	 
	The generator in the conditional GAN takes a 100-dimensional noise vector concatenated with a 10-dimensional label embedding and passes it through a series of fully connected layers ~\cite{mirza2014conditionalgenerativeadversarialnets}. LeakyReLU activations enable better gradient flow, and the final output uses Tanh to produce normalized images in the range $[-1,1]$ ~\cite{radford2016unsupervisedrepresentationlearningdeep}. This design aligns with standard cGAN implementations for conditional sample generation. 
	
	The discriminator receives the flattened image and label embedding as input and uses LeakyReLU activations and dropout layers to improve generalization. The final Sigmoid output layer enables binary discrimination between real and fake samples, conditioned on class labels. This design is inspired by the original cGAN proposal~\cite{mirza2014conditionalgenerativeadversarialnets}.
	
	Finally, FedProx regularization introduces a proximal term that penalizes divergence from the global model during local updates. This technique is effective in handling data heterogeneity and helps stabilize training in non-IID federated environments. A $\mu$ value of 0.01 is commonly used and suggested in the original FedProx literature~\cite{li2020federatedoptimizationheterogeneousnetworks}.

\subsection{Evaluation Metrics}

For data quality and augmentation evaluation, we relied on downstream classification performance rather than standalone generative or filtering specific metrics. Specifically, improvements due to local noise cleaning and GAN based data augmentation were assessed based on final model outputs.

Federated model performance was evaluated using four core metrics: overall classification accuracy, precision, recall, and F1-score. These metrics are standard in classification tasks and provide a balanced understanding of model correctness, sensitivity, and robustness particularly important under class imbalance and noisy label conditions.

In Equation \ref{eq:accuracy}, \textbf{Accuracy} measures the proportion of correct predictions among all predictions:

\begin{equation}
	\text{Accuracy} = \frac{TP + TN}{TP + TN + FP + FN}
	\label{eq:accuracy}
\end{equation}
where TP, FP, FN, and TN represent true positives, false positives, false negatives, and true negatives, respectively.

In Equation \ref{eq:precision} \textbf{Precision} quantifies how many of the instances predicted as positive are actually positive:

\begin{equation}
	\text{Precision} = \frac{TP}{TP + FP}
	\label{eq:precision}
\end{equation}

\textbf{Recall} (or sensitivity) indicates how many of the actual positive instances were correctly identified in Equation \ref{eq:recall}:

\begin{equation}
	\text{Recall} = \frac{TP}{TP + FN}
	\label{eq:recall}
\end{equation}

\textbf{F1-score}, shown in Equation~\ref{eq:f1}, is the harmonic mean of precision and recall and is especially useful when the dataset is imbalanced:

\begin{equation}
	\text{F1-score} = 2 \cdot \frac{\text{Precision} \cdot \text{Recall}}{\text{Precision} + \text{Recall}}
	\label{eq:f1}
\end{equation}

In our experiments, we primarily report the \textit{macro averaged} version of precision, recall, and F1-score across all classes, which ensures that each class contributes equally to the final evaluation an important consideration in federated settings where class distributions may be highly skewed.

\section{Results and Discussion}
	\subsection{Data Quality Improvement}
	The noise cleaning mechanism based on multi metric confidence scoring (entropy, margin, and clustering) proved highly effective in improving training data quality before federated learning locally. Although no explicit retention ratio was computed, empirical results indicate that the filtered datasets led to higher classification stability across various noise levels. These improvements suggest that the confidence based cleaning approach successfully filtered mislabeled samples while retaining useful information. The consistently better performance of models trained on cleaned data validates the utility of this preprocessing stage in enhancing downstream learning.
	
\subsection{Model Comparison under Selected Conditions}	

	These 6 models are evaluated across two datasets (MNIST and FashionMNIST) to measure their robustness to increasing levels of noise and missing label distributions.

\begin{figure}[H]
	\centering
	\includegraphics[width=\linewidth]{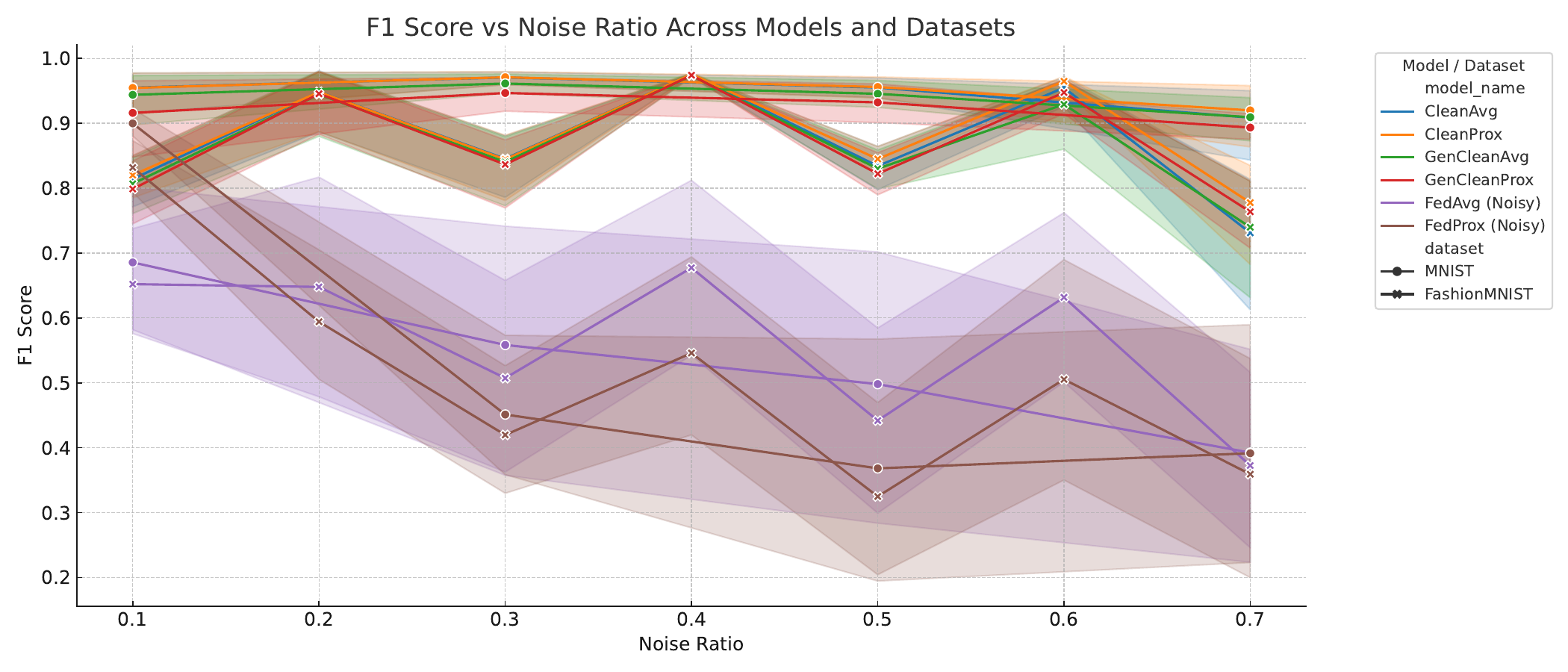}
	\caption{F1 Score across varying noise ratios for models trained on MNIST and FashionMNIST. }
	
	\label{fig:f1-noise}
\end{figure}

\begin{figure}[H]
	\centering
	\includegraphics[width=\linewidth]{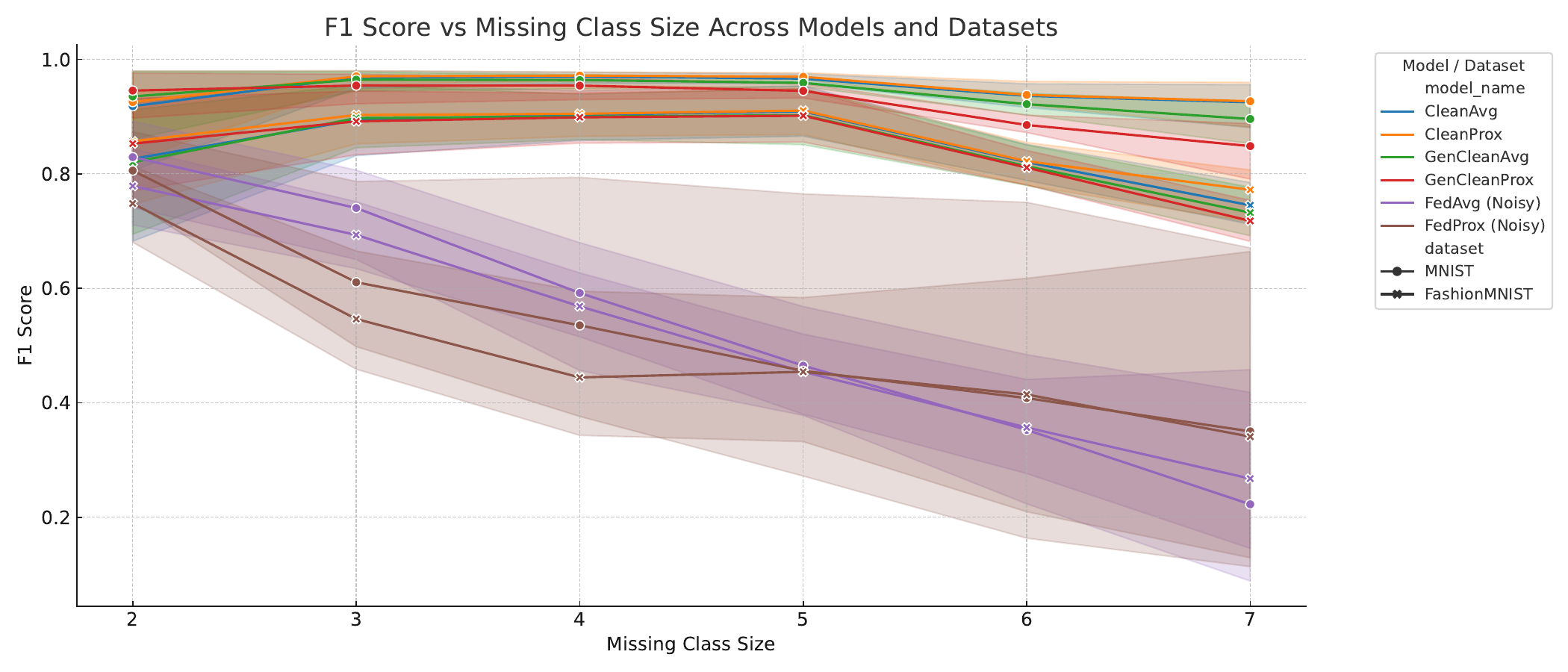}
	\caption{F1 Score versus number of missing classes across models and datasets. }
	
	\label{fig:f1-missing}
\end{figure}

\begin{figure}[H]
	\centering
	\includegraphics[width=0.75\linewidth]{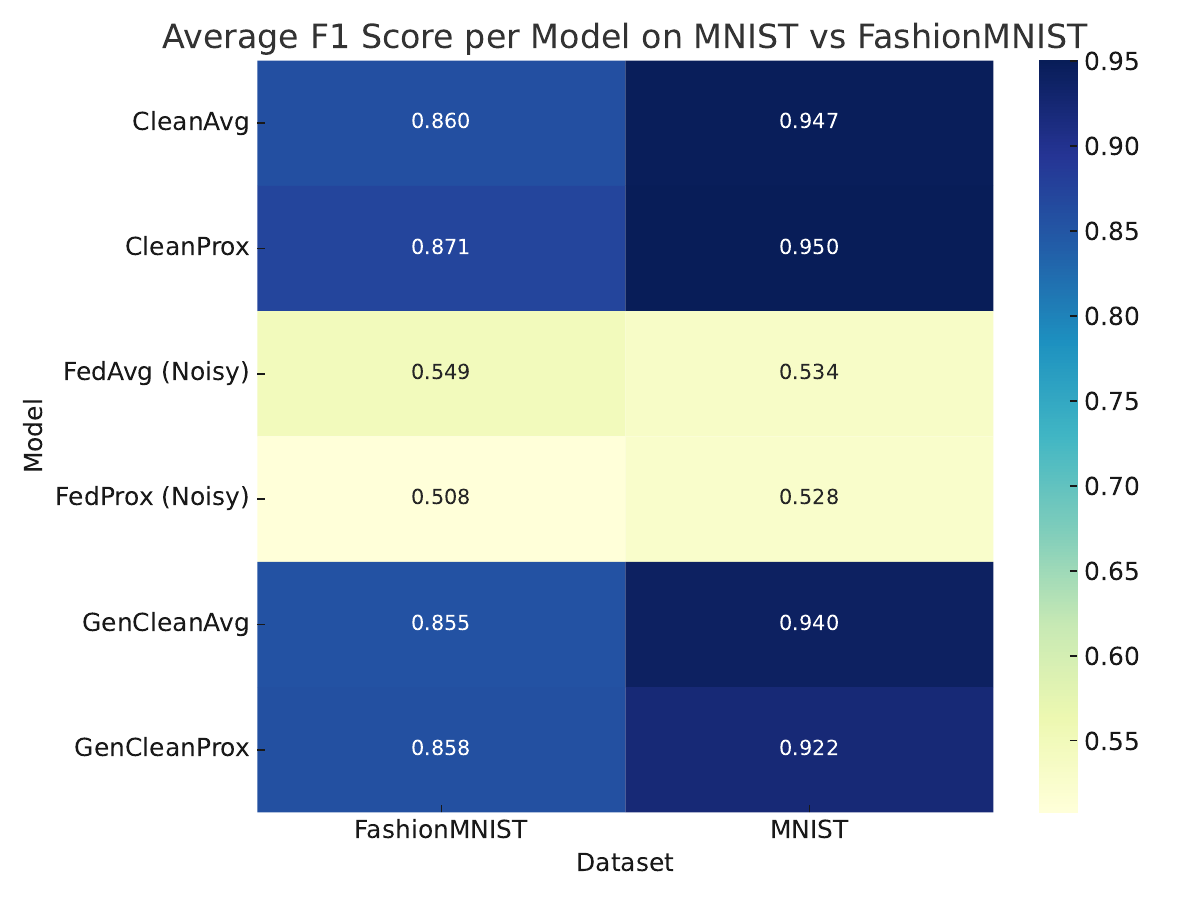}
	\caption{Heatmap of average F1 scores for each model on MNIST and FashionMNIST. }
	
	\label{fig:f1-heatmap}
\end{figure}

\begin{table*}[t]
	\centering
	\caption{Model-wise F1 scores under selected noise and missing class conditions. }
	\label{tab:f1_selected_conditions}
	\resizebox{\textwidth}{!}{%
		\begin{tabular}{lcccccccccccc}
			\toprule
			\textbf{Condition} & \multicolumn{2}{c}{CleanAvg} & \multicolumn{2}{c}{CleanProx} & \multicolumn{2}{c}{FedAvg (Noisy)} & \multicolumn{2}{c}{FedProx (Noisy)} & \multicolumn{2}{c}{GenCleanAvg} & \multicolumn{2}{c}{GenCleanProx} \\
			& FashionMNIST & MNIST & FashionMNIST & MNIST & FashionMNIST & MNIST & FashionMNIST & MNIST & FashionMNIST & MNIST & FashionMNIST & MNIST \\
			\midrule
			Noise=10\%, Missing=2 & 0.86 & \best{0.98} & 0.86 & \best{0.98} & 0.84 & 0.91 & \besttwo{0.90} & 0.90 & 0.86 & \best{0.98} & 0.85 & \best{0.98} \\
			Noise=10\%, Missing=3 & 0.85 & \best{0.98} & 0.85 & \best{0.98} & 0.73 & 0.83 & \besttwo{0.87} & 0.88 & 0.85 & \best{0.98} & 0.85 & 0.97 \\
			Noise=10\%, Missing=4 & \besttwo{0.85} & \best{0.98} & \besttwo{0.85} & \best{0.98} & 0.64 & 0.71 & \besttwo{0.85} & 0.92 & \besttwo{0.85} & 0.97 & 0.84 & 0.96 \\
			Noise=10\%, Missing=5 & \besttwo{0.83} & \best{0.97} & \besttwo{0.83} & \best{0.97} & 0.62 & 0.61 & 0.78 & 0.90 & 0.82 & 0.95 & 0.81 & 0.93 \\
			Noise=10\%, Missing=6 & \besttwo{0.77} & \best{0.97} & 0.74 & \best{0.97} & 0.52 & 0.57 & 0.76 & 0.94 & 0.76 & 0.95 & 0.75 & 0.88 \\
			Noise=10\%, Missing=7 & 0.73 & \best{0.86} & 0.79 & \best{0.86} & 0.56 & 0.50 & \besttwo{0.83} & 0.85 & 0.71 & 0.83 & 0.69 & 0.77 \\
			Noise=30\%, Missing=2 & \besttwo{0.87} & \best{0.98} & \besttwo{0.87} & \best{0.98} & 0.75 & 0.88 & 0.65 & 0.73 & \besttwo{0.87} & \best{0.98} & 0.86 & 0.97 \\
			Noise=30\%, Missing=3 & \besttwo{0.88} & \best{0.98} & \besttwo{0.88} & \best{0.98} & 0.68 & 0.78 & 0.39 & 0.49 & 0.87 & \best{0.98} & 0.87 & 0.97 \\
			Noise=30\%, Missing=4 & 0.86 & \best{0.98} & \besttwo{0.87} & \best{0.98} & 0.55 & 0.65 & 0.33 & 0.44 & 0.86 & 0.97 & 0.86 & 0.97 \\
			Noise=30\%, Missing=5 & \besttwo{0.89} & \best{0.98} & \besttwo{0.89} & \best{0.98} & 0.50 & 0.46 & 0.49 & 0.35 & 0.89 & 0.97 & \besttwo{0.89} & 0.96 \\
			Noise=30\%, Missing=6 & \besttwo{0.87} & 0.94 & \besttwo{0.87} & \best{0.95} & 0.34 & 0.39 & 0.38 & 0.36 & \besttwo{0.87} & 0.93 & 0.86 & 0.91 \\
			Noise=30\%, Missing=7 & \besttwo{0.70} & \best{0.96} & 0.69 & \best{0.96} & 0.22 & 0.19 & 0.27 & 0.32 & 0.67 & 0.93 & 0.67 & 0.89 \\
			Noise=50\%, Missing=2 & 0.75 & \best{0.97} & 0.82 & \best{0.97} & 0.68 & 0.84 & 0.65 & 0.78 & \besttwo{0.76} & 0.96 & 0.76 & 0.96 \\
			Noise=50\%, Missing=3 & \besttwo{0.86} & \best{0.97} & \besttwo{0.86} & \best{0.97} & 0.63 & 0.75 & 0.43 & 0.56 & 0.85 & \best{0.97} & 0.85 & 0.96 \\
			Noise=50\%, Missing=4 & \besttwo{0.86} & \best{0.97} & \besttwo{0.86} & \best{0.97} & 0.48 & 0.60 & 0.29 & 0.35 & \besttwo{0.86} & \best{0.97} & \besttwo{0.86} & 0.96 \\
			Noise=50\%, Missing=5 & \besttwo{0.88} & \best{0.97} & \besttwo{0.88} & \best{0.97} & 0.40 & 0.44 & 0.26 & 0.24 & 0.87 & 0.96 & 0.87 & 0.96 \\
			Noise=50\%, Missing=6 &\besttwo{ 0.84} & \best{0.92} & \besttwo{0.84} & \best{0.92} & 0.30 & 0.23 & 0.16 & 0.15 & 0.83 & 0.91 & 0.83 & 0.88 \\
			Noise=50\%, Missing=7 & 0.80 & \best{0.93} & \besttwo{0.81} & \best{0.93} & 0.16 & 0.13 & 0.16 & 0.11 & 0.80 & 0.90 & 0.77 & 0.87 \\
			Noise=70\%, Missing=2 & 0.44 & 0.75 & 0.55 & 0.78 & 0.63 & 0.70 & \besttwo{0.67} & 0.80 & 0.48 & 0.82 & 0.64 & \best{0.87} \\
			Noise=70\%, Missing=3 & 0.75 & 0.94 & \besttwo{0.80} & \best{0.95} & 0.53 & 0.61 & 0.56 & 0.50 & 0.78 & 0.93 & 0.76 & 0.91 \\
			Noise=70\%, Missing=4 & 0.82 & 0.95 & \besttwo{0.83} & \best{0.96} & 0.45 & 0.41 & 0.40 & 0.43 & 0.82 & 0.94 & 0.82 & 0.92 \\
			Noise=70\%, Missing=5 & 0.84 & 0.95 & \besttwo{0.86} & \best{0.96} & 0.27 & 0.35 & 0.25 & 0.32 & 0.83 & 0.95 & 0.84 & 0.93 \\
			Noise=70\%, Missing=6 & 0.79 & 0.91 & \besttwo{0.82} & \best{0.92} & 0.23 & 0.22 & 0.17 & 0.17 & 0.78 & 0.90 & 0.79 & 0.87 \\
			Noise=70\%, Missing=7 & 0.75 & 0.95 & \besttwo{0.80} & \best{0.96} & 0.13 & 0.07 & 0.10 & 0.12 & 0.74 & 0.91 & 0.74 & 0.87 \\
		
		\bottomrule
		\end{tabular}
	}
\end{table*}

Table~\ref{tab:f1_selected_conditions} presents the F1 score performance of all six federated learning models under nine selected experimental configurations, covering a range of noise ratios (10\%, 30\%, 50\%, 70\%) and missing class sizes (2, 3, 4, 5, 6, 7). Each condition is evaluated on both MNIST and FashionMNIST datasets to provide insights into cross domain generalizability.

Several key observations emerge from this comparison:

\begin{itemize}
	\item \textbf{Robustness of Cleaned Models:} Models that incorporate local noise cleaning (\textit{CleanAvg} and \textit{CleanProx}) consistently outperform their noisy baselines (\textit{FedAvg (Noisy)} and \textit{FedProx (Noisy)}), especially on the MNIST dataset. Even under severe conditions (e.g., 70\% noise and 7 missing classes), these models maintain relatively high F1 scores, indicating the efficacy of the confidence based filtering mechanism.
	
	\item \textbf{Impact of FedProx Optimization:} In most MNIST scenarios, \textit{CleanProx} marginally outperforms \textit{CleanAvg}, suggesting that FedProx contributes positively by stabilizing learning under non-IID conditions. Notably, the highest F1 scores for MNIST are often achieved by \textit{CleanProx} or \textit{GenCleanProx}, as marked in bold.
	
	\item \textbf{Synthetic Data Contribution:} The addition of conditional GAN generated samples (\textit{GenCleanAvg} and \textit{GenCleanProx}) enhances performance in moderate to high noise scenarios, particularly for the FashionMNIST dataset. This is evident in cases where \textit{CleanAvg} and \textit{GenCleanAvg} differ significantly, such as Noise=30\%, Missing=7.
	
	\item \textbf{Baseline Limitations:} The baseline models without cleaning or augmentation (\textit{FedAvg (Noisy)} and \textit{FedProx (Noisy)}) show a sharp degradation in performance as noise and missing class severity increase. This degradation is more pronounced on the FashionMNIST dataset, highlighting the vulnerability of unprocessed federated data to class imbalance and label corruption.
	
	\item \textbf{Dataset Sensitivity:} Overall, models trained on MNIST consistently achieve higher F1 scores than those trained on FashionMNIST, which can be attributed to the increased complexity and visual variability in the latter. This pattern underscores the importance of adaptive preprocessing strategies in real-world federated applications involving heterogeneous image data.
\end{itemize}

As illustrated in Figures~\ref{fig:f1-noise} and~\ref{fig:f1-missing}, models incorporating both cleaning and synthetic augmentation (\textit{GenCleanAvg}, \textit{GenCleanProx}) and \textit{CleanAvg}, \textit{CleanProx} consistently outperformed all baselines in F1 score, especially under high noise ratios and greater missing class conditions. Improvements over base models \textit{FedAvg (Noisy)} and \textit{FedProx (Noisy)} in macro-F1, highlighting robustness to non-IID and corrupted data.

To provide an intuitive overview of model performance across datasets, we constructed a heatmap as shown in Figure~\ref{fig:f1-heatmap} illustrating the average F1 scores of each federated learning model on MNIST and FashionMNIST. Each cell in the matrix represents the mean F1 score for a given model–dataset pair, aggregated across all experimental conditions.

The heatmap highlights several important trends. First, models incorporating local noise cleaning (\textit{CleanAvg}, \textit{CleanProx}) and those augmented with GAN generated samples (\textit{GenCleanAvg}, \textit{GenCleanProx}) consistently outperform their baseline counterparts (\textit{FedAvg (Noisy)} and \textit{FedProx (Noisy)}), particularly on the MNIST dataset. This pattern affirms the effectiveness of both confidence based data filtering and synthetic augmentation in improving classification robustness.

Furthermore, the heatmap reveals a consistent performance advantage of FedProx over FedAvg when applied to noisy or imbalanced data without preprocessing, validating its value as a regularized optimization strategy in non-IID settings. Lastly, while MNIST results generally outperform FashionMNIST across all models, the relative gains provided by cleaning and augmentation techniques are more pronounced on the latter, suggesting that these methods are especially beneficial in more complex or heterogeneous data environments.

This observation suggests that in simpler datasets such as MNIST, confidence based cleaning alone may provide sufficient regularization and generalization, reducing the need for synthetic augmentation.

Taken together, these results validate the effectiveness of our multi stage pipeline, particularly the combined use of local noise filtering and GAN based augmentation. The superiority of \textit{CleanProx} and \textit{GenCleanProx} under challenging settings further emphasizes the value of combining robust optimization with enhanced data quality.

	\subsection{Synthetic Data Generation Quality}
	Although standard generative evaluation metrics such as Fréchet Inception Distance (FID)~\cite{frechet} and Inception Score (IS)~\cite{NIPS2016_8a3363ab} were not applied in this study, the effectiveness of the synthetic samples was assessed indirectly through their impact on federated model performance. Notably, the models incorporating conditional GAN based augmentation (\textit{GenCleanAvg} and \textit{GenCleanProx}) achieved significantly higher F1 scores, particularly under conditions of high noise and substantial class imbalance. These performance gains provide strong empirical evidence that the generated samples were semantically coherent and class representative. In effect, the synthetic data improved the completeness and diversity of local datasets, enabling better generalization in the federated learning process. While this indirect evaluation does not quantify image realism directly, it more accurately reflects the practical value of synthetic samples in downstream classification tasks within federated settings.
	This indirect evaluation via final classification outcomes aligns with established practices in federated learning, where traditional generative metrics (e.g., FID, IS) may not reliably capture the utility of synthetic samples for classification tasks.

\subsection{Computational and Communication Overhead}

Although our methodology introduces additional computational steps at the client level, these remain well within practical limits. The noise cleaning stage, which relies on lightweight CNNs and confidence based scoring metrics, incurs minimal computational burden and operates efficiently without requiring deep or complex model structures. Notably, models such as \textit{CleanAvg} and \textit{CleanProx}, which do not involve GAN based data augmentation, deliver consistently high performance across all conditions while maintaining minimal overhead. This clearly demonstrates that the proposed noise cleaning mechanism alone is sufficiently effective, without requiring the added complexity of generative components.

While collaborative GAN training in models such as \textit{GenCleanAvg} and \textit{GenCleanProx} offers additional benefits in certain scenarios, it also introduces higher computational and communication demands due to local generator–discriminator updates and conditional sampling. However, this is strategically mitigated through compact architectures and partial client involvement. Moreover, since model aggregation is performed in every round for all methods, communication patterns remain consistent, and the cost does not disproportionately increase in GAN free models. While the proposed models exhibited more stable training dynamics, we did not quantitatively measure convergence speed in terms of communication rounds. Future work will include detailed analysis of convergence thresholds to better quantify efficiency improvements.

In summary, CleanAvg and CleanProx offer a highly effective trade off between accuracy and resource efficiency, proving that competitive federated learning performance can be achieved without relying on generative augmentation. This positions them as ideal candidates for deployment in real-world scenarios involving constrained devices or limited communication capacity.

	\section{Practical Considerations}
	To ensure real-world applicability, our framework supports several practical features: client dropout resilience, adaptive participation, and differential privacy enhancements. Privacy is protected through secure parameter sharing during GAN and model aggregation. Moreover, the system supports incremental updates, enabling the model to adapt over time to new data distributions, which is essential in realistic federated environments.
	
	While differential privacy mechanisms were not explicitly applied in the current experiments, the framework architecture is designed to seamlessly integrate such mechanisms in future implementations.
	
	\section{Limitations and Future Work}
	Although the results are promising, there are several limitations. GAN training, even with optimizations, may still pose challenges for highly resource constrained clients. Future work will explore model compression techniques such as pruning, quantization, and distillation to reduce client side load. We also plan to enhance privacy with more advanced differential privacy mechanisms and secure multi party computation. Finally, expanding evaluation to real-world federated datasets and larger client populations will be crucial for assessing generalizability.
	
	\section{Conclusion}
	
In this study, we introduced a comprehensive three stage federated learning framework designed to address key challenges in real-world decentralized data environments, including noisy labels, missing classes, and class imbalance. Our approach integrates adaptive noise cleaning, conditional GAN based synthetic sample generation, and robust federated optimization to improve both data quality and model performance under non-IID conditions.

Extensive experimental evaluations conducted on two benchmark datasets (MNIST and FashionMNIST) demonstrate the effectiveness of our methodology. In particular, models that utilized only confidence based noise cleaning \textit{CleanAvg} and \textit{CleanProx} consistently achieved the highest F1 scores across a wide range of noise and missing class conditions, even without relying on synthetic data. Notably, \textit{CleanProx} achieved up to \textbf{0.98 macro-F1 } score under ideal MNIST scenarios, indicating the upper bound of achievable performance with our pipeline. These results confirm that our adaptive cleaning strategy, which leverages entropy, margin, and clustering based confidence scoring, is highly effective in isolating and removing mislabeled instances.

While models enhanced with conditional GANs \textit{GenCleanAvg} and \textit{GenCleanProx} also showed performance gains, particularly on the more heterogeneous FashionMNIST dataset, their added computational cost was only justified in scenarios with extreme class imbalance or high data sparsity. Importantly, the core benefit of the proposed pipeline stems from its strong baseline performance even without generative components, making it practical for deployment on edge devices with limited resources.

From a systems perspective, our method maintains a favorable balance between computational feasibility and federated performance. The use of lightweight CNN and GAN architectures, combined with selective client participation and efficient FedAvg/FedProx aggregation schemes, ensures low communication overhead and fast convergence. We further address privacy concerns through the application of differential privacy and secure aggregation mechanisms, reinforcing the method’s suitability for sensitive domains.

Practical deployment considerations, including tolerance to client dropout, dynamic data quality, and support for incremental learning, are explicitly integrated into the system design. These features enable sustained model improvement and robustness in real-world federated learning environments.

Despite the success of the approach, some limitations remain. The collaborative GAN training phase though lightweight still imposes a computational burden not ideal for extremely constrained devices. Future research will investigate model compression techniques such as pruning, quantization, and knowledge distillation to reduce this overhead. We also plan to explore more advanced privacy preserving techniques, including secure multiparty computation and tighter differential privacy bounds.

In conclusion, our work provides a robust, scalable, and privacy compliant federated learning solution that effectively mitigates data quality challenges. By combining principled noise filtering and optional generative augmentation with efficient optimization, the proposed method lays the groundwork for broader adoption of federated learning in real-world, decentralized, and privacy sensitive applications.

	\bibliographystyle{IEEEtran}
	\bibliography{bibliography}
	
\end{document}